%% file: main.tex
\documentclass{article}

\usepackage[preprint]{neurips_2026}
\usepackage{amsthm}
\newtheorem{definition}{Definition}
\usepackage{booktabs}
\usepackage{xcolor}
\usepackage[table]{xcolor}
\newcommand{\cmark}{{\color{green!60!black}\checkmark}}
\newcommand{\xmark}{{\color{red!60!black}\texttimes}}
\newcommand{\pmark}{{\color{orange!80!black}$\sim$}}
\usepackage{graphicx}
\usepackage{enumitem}
\usepackage{hyperref}
\usepackage{subcaption}
\usepackage{amsmath}
\usepackage[capitalize]{cleveref}

\usepackage[utf8]{inputenc} 
\usepackage[T1]{fontenc}    
\usepackage{hyperref}       
\usepackage{url}            
\usepackage{xurl}           
\usepackage{booktabs}       
\usepackage{amsfonts}       
\usepackage{nicefrac}       
\usepackage{microtype}      
\usepackage{xcolor}         

\title{QUIVER: A Formal Framework for Quantifying Perturbation Propagation and Bifurcation in Compound AI Systems}

\author{%
    Prashanti Nilayam \\
    Servicenow\\
    CA, USA \\
    \And
    Sankalp Nayak \\
    Servicenow \\
    CA, USA \\
  }

\begin{document}
\maketitle
\input{abstract-compressed}
\begin{figure}[t]
  \centering
  \begin{subfigure}[b]{0.48\textwidth}
    \centering
    \includegraphics[width=\linewidth]{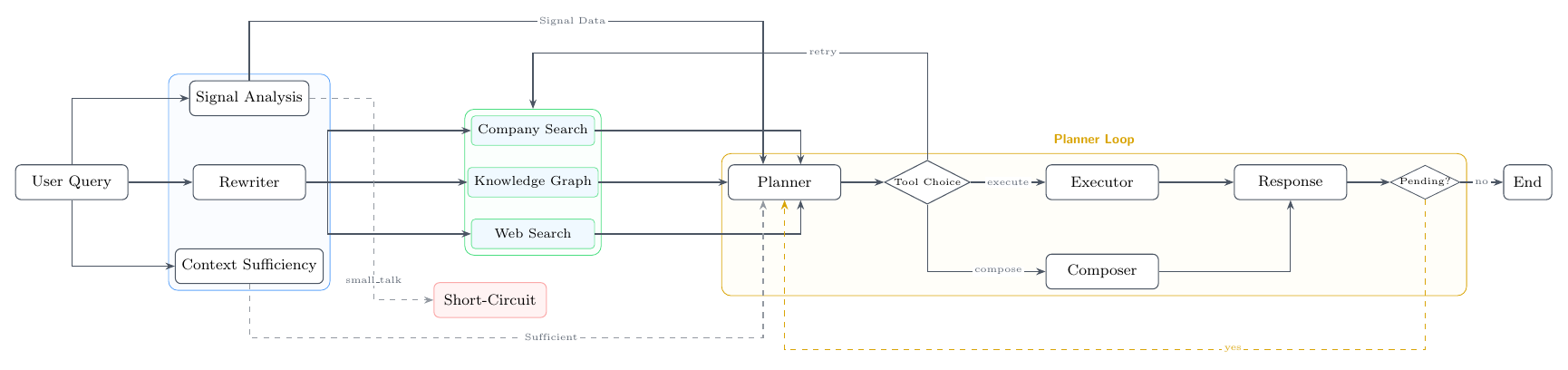}
    \caption{System P architecture.}
    \label{subfig:sys_p}
  \end{subfigure}
  \hfill
  \begin{subfigure}[b]{0.48\textwidth}
    \centering
    \includegraphics[width=\linewidth]{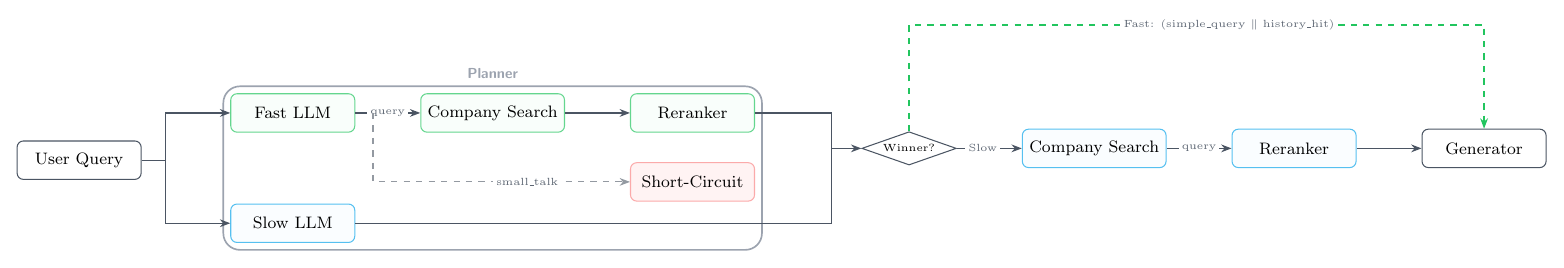}
    \caption{System Q architecture.}
    \label{subfig:sys_q}
  \end{subfigure}

  \caption{\textbf{Comparative architectures of Systems~P and~Q.} System~P uses a parallel-intake first wave (rewriter, signal analysis, context sufficiency) feeding a planner loop with conditional replanning and tool execution. System~Q uses a dual-planner entry point with a winner-selection mechanism that routes to either a fast path or a slow path through retrieval, reranking, and generation.}
  \label{fig:comparison}
\end{figure}

\input{intro1}
\input{framework}
\input{related-work}
\input{experimentsetup1}
\input{results}
\input{limitations}

\bibliographystyle{abbrvnat}
\bibliography{references}
\include{appendix1}

\end{document}

%% file: abstract-compressed.tex
\begin{abstract}
Compound AI systems that chain multiple LLM calls into directed computation graphs with parallel branches, sequential stages, and conditional loops, are now the dominant architecture for production AI applications ~\citep{khattab2024dspy, yao2022react, shao2024assisting}. Although these architectures leverage heterogeneous nodes with mixed-mode outputs, from typed schemas to natural language, no existing framework quantifies how perturbations propagate through such pipelines, where nodes are stochastic, outputs are heterogeneously typed, and execution paths can diverge structurally under perturbation. We introduce \textbf{QUIVER}, a formal framework for measuring perturbation propagation in computation-graph-structured LLM pipelines. The framework defines: (1) a sensitivity matrix with type-dispatched distance metrics that classifies edges as amplifiers, absorbers, or threshold-sensitive and composes multiplicatively along paths, complemented by occurrence-lift capturing the probability of downstream drift independent of magnitude; (2) three-component trajectory divergence decomposing variation into value drift, structural path divergence, and iteration count divergence; (3) bifurcation thresholds identifying the smallest perturbation that causes structural execution path changes; and (4) distribution faithfulness, quantifying when per-node evaluation datasets diverge from production distributions. We validate on two production enterprise pipelines with different architectures (Systems~P and~Q; Figures \ref{subfig:sys_p}, \ref{subfig:sys_q}) and cross-validate on a public DSPy \cite{khattab2024dspy} multi-hop QA pipeline (HotpotQA ~\cite{yang2018hotpotqa}/ColBERTv2~\cite{santhanam2022colbertv2}), a third and structurally distinct topology. Across 8,200+ instrumented traces yielding 32,000+ pair comparisons, we demonstrate that: (a)~the framework reveals distinct sensitivity profiles across architectures, from global resilience to deep cascade amplification; (b)~identical divergence rates arise from mechanistically distinct patterns, distinguishable only via nodal decomposition; (c)~observational sensitivity profiles predict nodes prone to trajectory bifurcation; and (d)~distribution faithfulness quantifies per-node evaluation gap, localizing stale-evaluation artifacts to specific node-field categories that aggregate metrics cannot surface (Appendix~F case study).
\end{abstract}

%% file: intro1.tex
\section{Introduction}

Compound AI systems that chain multiple LLM calls into directed computation graphs with parallel branches, sequential stages, and conditional loops are now the dominant architecture for production AI applications~\citep{khattab2024dspy, yao2022react, shao2024assisting}. Although these architectures span heterogeneous nodes with mixed-mode outputs [typed schemas to natural language], no existing framework quantifies how perturbations propagate through them, where nodes are stochastic, outputs are heterogeneously typed, and execution paths can diverge structurally under perturbation.

When a production pipeline returns a low-quality response, practitioners cannot determine which node is responsible, whether quality degraded continuously along a chain of small perturbations, or whether a single upstream change crossed a threshold that flipped the execution path. End-to-end evaluation detects symptoms but cannot localize causes; per-node evaluation in isolation tests each component against curated inputs that may not reflect what it receives from upstream, yielding quality estimates that do not predict production behavior. The problem compounds in pipelines with conditional loops~\citep{yao2022react}, where small upstream perturbations can activate different nodes, execute different loop iterations, or retrieve entirely different knowledge. Existing optimization and evaluation tooling~\citep{khattab2024dspy, yuksekgonul2024textgrad, cheng2024trace} treats nodes independently, without measuring how perturbations propagate across edges or where structural bifurcation occurs (Section~3).

We introduce \textbf{QUIVER}, a formal framework for measuring perturbation propagation in computation-graph-structured LLM pipelines. The framework defines four contributions: (1)~a \emph{sensitivity matrix} with type-dispatched distance metrics that classifies edges as amplifiers, absorbers, or threshold-sensitive, complemented by \emph{occurrence-lift} capturing the probability of propagation independent of magnitude; (2)~\emph{three-component trajectory divergence} decomposing variation into value drift, structural path divergence, and iteration count divergence; (3)~\emph{bifurcation thresholds} identifying the smallest perturbation that causes structural execution path changes; and (4)~\emph{distribution faithfulness}, quantifying when evaluation datasets diverge from production distributions. We also release a programmatic trace interface for automatic computation of all measurements on any pipeline.

We validate on two production enterprise pipelines with different architectures (Figure~\ref{fig:comparison}): System~P, a complex graph with parallel intake, retrieval tool selection, and a conditional replanning loop ($k_{\max}{=}5$); and System~Q, a dual-planner pipeline with parallel reranking and fast/slow path routing. We additionally cross-validate on a public DSPy multi-hop QA pipeline over HotpotQA with ColBERTv2 retrieval which is a strict sequential chain that supplies a third, structurally distinct topology. Across 8{,}200+ instrumented traces yielding 32{,}000+ pair comparisons (observational and interventional), we demonstrate: (a)~the framework reveals distinct sensitivity profiles across architectures, from global resilience to deep cascade amplification; (b)~identical divergence rates arise from mechanistically distinct cascade patterns, distinguishable only via per-node decomposition; (c)~observational profiles predict which nodes will bifurcate under perturbation; and (d)~distribution faithfulness detects previously undetectable evaluation invalidation from configuration changes.

%% file: framework.tex
\section{Framework}

We define the formal objects for measuring perturbation propagation in compound LLM pipelines. Extended discussion, worked examples, and estimation details are provided in Appendices~A and~B.

\subsection{Pipeline Graph and Typed Output Spaces}

\begin{definition}[Typed Pipeline Graph]
A compound LLM pipeline is a tuple $\mathcal{G} = (V, E, \mathcal{T}, \mathcal{F})$ where $V = \{v_1, \ldots, v_n\}$ is a finite set of nodes, $E \subseteq V \times V$ is a set of directed edges representing data flow, $\mathcal{T} = \{T_1, \ldots, T_n\}$ assigns each node $v_i$ a typed output space $T_i$, and $\mathcal{F} = \{f_1, \ldots, f_n\}$ assigns each node a stochastic function:
\begin{equation}
f_i : \prod_{j \in \mathrm{pa}(i)} T_j \rightarrow \Delta(T_i)
\end{equation}
where $\mathrm{pa}(i) = \{j : (v_j, v_i) \in E\}$ and $\Delta(T_i)$ is the set of probability distributions over $T_i$.
\end{definition}
\noindent Typed output spaces may include schema-typed objects, ordered lists, categorical values, or unstructured text. External inputs are modeled as source nodes with no parents.

\subsection{Type-Dispatched Distance Metrics}

\begin{definition}[Type-Dispatched Distance]
For each typed output space $T_i$, define $d_i : T_i \times T_i \rightarrow \mathbb{R}_{\geq 0}$ satisfying non-negativity and identity of indiscernibles. For schema-typed spaces $T_i = T_i^{(1)} \times \cdots \times T_i^{(m)}$:
\begin{equation}
d_i(x, y) = \sum_{k=1}^{m} w_k \cdot d_i^{(k)}(x^{(k)}, y^{(k)})
\end{equation}
where $w_k \geq 0$, $\sum_k w_k = 1$, and each $d_i^{(k)}$ is appropriate to the field type: $\mathbb{1}[a \neq b]$ for categorical, $1 - |A \cap B|/|A \cup B|$ for set-valued, normalized edit distance for ordered lists, normalized absolute difference for numeric, and $1 - \cos(\phi(s), \phi(t))$ for text fields.
\end{definition}
\noindent Per-field weights $w_k$ may encode application priors (e.g., higher weight for routing-decision fields than for descriptive context fields); the choice of weighting and per-type kernel is application-dependent and orthogonal to the framework's edge-classification definitions below.

\subsection{Sensitivity Matrix}

\begin{definition}[Edge Sensitivity]
For each edge $(v_i, v_j) \in E$:
\begin{equation}
\sigma_{ij} = \mathbb{E}\left[\frac{d_j\big(f_j(\mathbf{x}),\; f_j(\mathbf{x}')\big)}{d_i(x_i, x_i')}\right]
\end{equation}
where $\mathbf{x}, \mathbf{x}'$ differ only in the component from $v_i$, and the expectation is restricted to pairs with $d_i > \epsilon$.
\end{definition}
\noindent Classifies edges as \emph{amplifiers} ($\sigma > 1$), \emph{absorbers} ($\sigma < 1$), or \emph{insensitive} ($\sigma \approx 0$). Edges with $\hat{\sigma}$ close to 1 are also called \emph{near-unity}: they sit at the amplifier/absorber boundary, where the binary classification is most sensitive to per-field kernel and weight choices. The band $|\hat{\sigma}-1| < \delta$ for an application-chosen $\delta$ captures these edges; we report $\delta$ inline where used. The scalar $\sigma_{ij}$ may summarize a multimodal distribution; we recommend inspecting the full ratio distribution (Section~5).

\begin{definition}[Occurrence-Lift]
For each edge $(v_i, v_j) \in E$:
\begin{equation}
\lambda_{ij} = P\big(d_j > 0 \mid d_i > 0\big) - P\big(d_j > 0 \mid d_i = 0\big)
\end{equation}
\end{definition}
\noindent Captures the \emph{probability} of downstream drift, complementing $\sigma_{ij}$ which captures \emph{magnitude}. The two measures are decoupled, carrying distinct information about an edge (Section~5).

\begin{definition}[Sensitivity Matrix]
$\Sigma \in \mathbb{R}_{\geq 0}^{n \times n}$ where $\Sigma_{ij} = \sigma_{ij}$ if $(v_i, v_j) \in E$, else $0$.
\end{definition}

\begin{definition}[Path Sensitivity]
For a directed path $p = (v_{i_1}, \ldots, v_{i_k})$:
\begin{equation}
\sigma(p) = \prod_{(v_{i_l}, v_{i_{l+1}}) \in p} \sigma_{i_l, i_{l+1}}
\end{equation}
\end{definition}
\noindent A path is a \emph{cascade amplifier} if $\sigma(p) > 1$. The maximum over all source-to-sink paths is the \emph{critical amplification path}. When parallel branches reconverge, interaction terms may be estimated via multivariate regression (Appendix~B).

\subsection{Loop-Augmented Pipelines and Trajectory Bifurcation}

We extend the framework to pipelines with conditional loops~\citep{yao2022react}. The loop body $L \subseteq V$ is unrolled into indexed copies $L^{(1)}, \ldots, L^{(k^*)}$ where $k^* \leq k_{\max}$.

\begin{definition}[Iteration Action]
At each iteration $t$, the loop controller produces $a^{(t)} \in \mathcal{A}$ from a finite action set (e.g., \texttt{EXECUTE}, \texttt{RETRY}, \texttt{GENERATE}, \texttt{COMPOSE}) with auxiliary parameters $Q^{(t)}$. The iteration topology is $g^{(t)} = (a^{(t)}, Q^{(t)})$; the trajectory topology is $G^* = (k^*, g^{(1)}, \ldots, g^{(k^*)})$.
\end{definition}

\begin{definition}[Trajectory]
$\tau = (o_1, \ldots, o_n, o_L^{(1)}, \ldots, o_L^{(k^*)})$ where $o_i \in T_i$ and $o_L^{(t)}$ is the output tuple at iteration $t$.
\end{definition}

\begin{definition}[Trajectory Divergence]
For trajectories $\tau, \tau'$:
\begin{equation}
D(\tau, \tau') = (D_{\mathrm{iter}},\; D_{\mathrm{shape}},\; D_{\mathrm{output}})
\end{equation}
where $D_{\mathrm{iter}} = \sum_{i} |c_i - c_i'|$ with $c_i$ the number of invocations of node $v_i$ in trajectory $\tau$ (this generalizes loop-iteration counting: for a loop body $L$, $\sum_{i \in L} |c_i - c_i'|$ recovers $|L|\cdot|k^* - k^{*'}|$ when iteration shapes match, and otherwise also captures sub-stage call-count divergence in non-loop pipelines such as reranker or tool-execution invocations), \; $D_{\mathrm{shape}} = \sum_{t=1}^{\min(k^*, k^{*'})} \mathbb{1}[g^{(t)} \neq g'^{(t)}]$, \; and $D_{\mathrm{output}} = \sum_i w_i \cdot d_i(o_i, o_i')$.
\end{definition}
\noindent The three components capture per-node \emph{count}, \emph{order}, and \emph{value} divergence respectively. We additionally derive a node-presence indicator $D_{\mathrm{struct}}(\tau, \tau') := \mathbb{1}[\{v : v \in \tau\} \neq \{v : v \in \tau'\}]$ from the trajectories as a summary statistic; it is not a fourth component of $D$ but is reported alongside it because $D_{\mathrm{iter}}$ and $D_{\mathrm{shape}}$ alone do not flag whether the activated node sets differ. The indicators may co-occur. For loop-free pipelines this construction extends naturally: we take $k^* = 1$ and define $g^{(1)} = (a^{(1)}, Q^{(1)})$ as the trace's \emph{conditional-branch activation vector} --- the realized values of all branching/routing decisions in $\tau$ (which arms of conditional gates were taken, which optional nodes activated). $D_{\mathrm{shape}} = \mathbb{1}[g^{(1)} \neq g'^{(1)}]$ then compares branch activations across $\tau, \tau'$. We treat this as a natural extension of Definition~7 rather than an additional formal construct; the loop and non-loop cases use the same indicator on $g^{(t)}$.

\begin{definition}[Bifurcation Threshold]
For node $v_i$ upstream of a loop or conditional branch:
\begin{equation}
\beta_{\mathrm{shape}}(v_i) = \inf\{d_i(f_i(x), f_i(x')) : D_{\mathrm{shape}} > 0\}
\end{equation}
$\beta_{\mathrm{iter}}(v_i)$ is defined analogously for $D_{\mathrm{iter}} > 0$.
\end{definition}
\noindent Nodes with small $\beta_{\mathrm{shape}}$ are \emph{bifurcation-sensitive}: minor output variations cause structural path divergence that $\sigma_{ij}$ alone cannot detect.

\paragraph{Noise origins.} Node $v_i$ is a \emph{noise origin} if $P(d_i > \epsilon \mid d_j = 0\;\forall\, j \in \mathrm{pa}(i)) > 0$; otherwise it is a \emph{propagator}. This partitions nodes into intrinsic variance sources and inherited variance carriers.

\subsection{Evaluation Principles and Estimation}

\paragraph{Distribution Faithfulness.} A per-node evaluation is \emph{distribution-faithful} w.r.t.\ edge $(v_i, v_j)$ if $D_{\mathrm{KL}}(P_{\mathrm{prod}}(T_i) \,\|\, P_{\mathrm{eval}}(T_i)) < \delta$. Unfaithful evaluation at high-$\sigma$ edges amplifies quality estimation errors downstream.

\paragraph{Cross-Stage Regression Detection.} The \emph{impact set} $\mathcal{I}(v_i, \alpha) = \{v_j : \sigma(p) > \alpha \text{ for some } p\colon v_i\to v_j\}$ identifies downstream nodes requiring re-evaluation when $v_i$ changes; per-edge \emph{drift budgets} $\tau_{ij}$ specify the maximum upstream drift below which $v_j$ remains within its noise floor.

\paragraph{Estimation.} $\sigma_{ij}$, $\lambda_{ij}$, $\tau_{ij}$, and noise origin classification are estimable from \emph{observational} traces via partial regression; bifurcation thresholds $\beta_{\mathrm{shape}}, \beta_{\mathrm{iter}}$ require \emph{interventional} experiments because they describe decision boundaries rather than continuous relationships. Definition~3 is stated interventionally; the observational estimator approximates it by partialing out parent co-variation, which is unbiased only when parents have no shared unmeasured causes. When intake nodes share an upstream input (e.g., a common user query in System~P), residual confounding can bias $\hat{\sigma}_{ij}$; we use the interventional corpus (Section~5.2) to validate the observational estimates at perturbation-targetable edges, and treat agreement between the two as evidence that the approximation holds for the pipelines studied. Observational identifies proximate bifurcation sources; interventional reveals dormant risks beyond the natural operating envelope. The matrix is locally valid at the current operating point and should be re-estimated after large configuration changes (\S\ref{sec:limitations}). Full details: Appendix~B.

%% file: related-work.tex
\section{Related Work}

\paragraph{Optimization of compound AI systems.} DSPy~\citep{khattab2024dspy} introduces typed signatures and optimizes end-to-end through prompt tuning; TextGrad~\citep{yuksekgonul2024textgrad} backpropagates textual feedback as gradients; Trace/OPTO~\citep{cheng2024trace} and LLM-AutoDiff~\citep{yin2025llm} extend this with richer execution traces. \citet{chen2026textual} shows textual gradients degrade exponentially with depth, and TextResNet~\citep{huang2026textresnet} identifies semantic entanglement as a deep-pipeline challenge, both respond with better optimizers. The prior step is missing: practitioners need to understand propagation dynamics before optimizing. The EMNLP~2025 survey~\citep{lee2025compound} reviews 26 optimization methods without identifying sensitivity analysis as a direction.

\paragraph{Failure attribution in agentic systems.} AgenTracer~\citep{zhang2025agentracer} replaces each agent action with an oracle-corrected alternative to identify the earliest fault-flipping correction; RAFFLES~\citep{zhu2026raffles} uses LLM-based causal-hierarchy reasoning. Both produce binary attribution rather than continuous sensitivity measurement, and both require auxiliary models that introduce uncharacterized error. Our framework requires no auxiliary model: measurements are computed directly from production traces via type-dispatched distance and standard regression. The text-distance kernel takes any sentence-embedding $\phi$ (we use \texttt{all-MiniLM-L6-v2}); $\phi$ is non-generative,  not an LLM judge or oracle, so the ``no auxiliary model'' claim specifically excludes generative/judge models, the distinction that matters for evaluation cost and reliability. Watershed~\citep{watershed2025eval} distinguishes property from correctness evaluations at component and task levels; our work formalizes this with the sensitivity matrix and distribution faithfulness~\citep{es2024ragas}.

\paragraph{Formal analysis of LLM system dynamics.} UProp~\citep{duan2025uprop} decomposes per-step uncertainty into intrinsic and extrinsic components via pointwise mutual information on sequential chains, requiring model logprobs; we handle arbitrary computation graphs at the typed-output level using only logged intermediate outputs. ``From Spark to Fire'' \citep{xie2026spark} formalizes cascade amplification via spectral-radius analysis but assumes homogeneous interaction matrices; our typed representations require heterogeneous, type-dispatched per-edge metrics~\citep{shen2025understanding}. \citet{he2025information} apply information-bottleneck analysis, showing stage-wise information loss is unrecoverable. ``Geometric Dynamics''~\citep{tacheny2025dynamics} classifies loop behavior as contractive/oscillatory/exploratory but does not address upstream-perturbation-driven transitions, our bifurcation thresholds do. \citet{kim2025towards} measure a $17.2\times$ error amplification in unstructured multi-agent networks, giving empirical context for our per-edge cascade quantification~\citep{liu2026valueflow,eslami2026control}.

No existing approach combines computation-graph-level sensitivity analysis, type-dispatched distance, bifurcation analysis for conditional paths, and the distribution-faithfulness / cross-stage-regression evaluation principles (capability comparison in Appendix~\ref{app:capability}).


%% file: experimentsetup1.tex
\section{Experimental Setup}

We validate the framework on two production enterprise conversational AI pipelines with different architectures, referred to as System~P and System~Q (Figure~\ref{fig:comparison}).

\paragraph{System P} is a three-phase computation graph: a \emph{First Wave} of parallel intake nodes (rewriter, signal analysis, context sufficiency); a \emph{Discovery} phase where the planner selects and merges retrieval tools (company search, knowledge graph, web search); and a \emph{Planner Loop} ($k_{\max}=5$) that iterates over subtasks, choosing per iteration to execute a tool, compose, or retry discovery. Signal analysis gates conditional paths (small-talk short-circuit, web-search-hint).

\paragraph{System Q} is a dual-planner pipeline: a fast and slow LLM process the query in parallel, and a winner-selection mechanism routes to a fast path (simple queries, history cache hits) or a slow path through company search, reranking, and generation. It has no replanning loop but branches conditionally at the planner and fast/slow selection.

\paragraph{Trace collection.} LLM calls use a fixed determinism contract (temperature${}=0$, seed${}=42$) via Azure OpenAI; text distances use \texttt{all-MiniLM-L6-v2}. System~P: an \emph{observational corpus} of 1{,}500 traces (500 seeds $\times$ 3 repeats) and an \emph{interventional corpus} of 3{,}183 baseline-perturbed pairs across five classes (rewriter swap, signal turn-type/web-search-hint/small-talk flips, context-sufficiency override). System~Q: 1{,}497 observational traces drawn from production replay across 209 distinct utterance payloads with \emph{heterogeneous} per-utterance group sizes (production traffic is not uniform; some utterances appear far more often than others). Pairs are enumerated as all unordered same-input pairs within each group, $\sum_g \binom{n_g}{2} = 24{,}693$. The pair count substantially exceeds the $500\cdot\binom{3}{2}{=}1{,}500$ implied by a balanced 500$\times$3 design because of the skewed group-size distribution.

\paragraph{Perturbation design.} System~P perturbations target upstream nodes at controlled magnitudes: rewriter swaps four prompt variants (mean $d_{\mathrm{rewriter}} \in [0.696, 0.754]$); signal analysis flips categorical fields (\texttt{turn\_type}, \texttt{web\_search\_hint}, \texttt{is\_small\_talk}); context sufficiency forces $\texttt{sufficient}{=}\texttt{True}$, removing discovery from the graph. Interventional results are stratified by effective vs.\ no-op pairs (the latter, where baseline already matches the perturbed value, serve as built-in negative controls).

\paragraph{Distance metrics.} We use uniform per-field aggregation under the type-dispatched kernels of Definition~2. Definition~2 admits any per-field weights $w_k \geq 0$ with $\sum_k w_k = 1$ and any per-type kernel; the specific instantiation is application-dependent and the framework's measurements ($\hat{\sigma}$, $\lambda$, $\beta$) are well-defined for any admissible choice.

%% file: results.tex
\section{Results}

We validate each formal object from Section~2 on production traces from Systems~P and~Q. Our goal is not to characterize these specific pipelines but to demonstrate that the framework's measurements are computable, reveal non-trivial structure, and generalize across architectures.

\subsection{Sensitivity Coefficients Reveal Threshold-Sensitive Structure}
\label{sec:sensitivity}

Table~\ref{tab:sigma} reports edge sensitivity $\hat{\sigma}_{ij}$ and occurrence-lift $\lambda_{ij}$ for System~P: two edges amplify ($\hat{\sigma} > 1$; rewriter$\to$discovery, planner$\to$composer), four absorb. The full ratio distribution reveals that mean-classified amplifiers are bimodal,  a majority absorber regime (52--65\% of pairs, ratio ${<}1$) and a minority amplification tail (12--23\%, ratio ${>}1.5$, max ${\sim}7\times$); the mean lands near unity because the two regimes roughly balance. Tail pairs cluster around two mechanisms: retrieval-boundary flips at single-word query changes, and categorical field flips at composition (the latter contributes minimally to composite output distance but gates execution routing).

$\hat{\sigma}$ and $\lambda$ are decoupled, carrying distinct information (Table~\ref{tab:sigma}): rewriter$\to$discovery has modest $\hat{\sigma}{=}1.128$ but near-total coupling $\lambda{=}{+}0.994$; signal$\to$planner has meaningful $\hat{\sigma}{=}0.857$ but negligible $\lambda{=}{+}0.041$.

System~Q exhibits a qualitatively different sensitivity profile (Table~\ref{tab:sigma_q}). Multiple edges show $\hat{\sigma} \gg 1$, with amplification factors up to $9\times$ on well-sampled edges. Fast~LLM and Slow~LLM (the parallel intake planners) exhibit moderate $D_{\mathrm{noise}}$ (0.180, 0.172) consistent with their first-stage position as likely intrinsic variance sources; the formal noise-origin classification (\S2.4, Appendix~B.4) is not reported for System~Q because the production-replay corpus lacks pairs with byte-identical upstream payloads at every intake node simultaneously. The framework correctly identifies the post-Reranker absorber edge ($\hat{\sigma}=0.34$) and the deterministic tool-execution terminals (insensitive, $\hat{\sigma}=0$ from all upstream nodes).

\begin{table}[t]
\caption{\textbf{Edge sensitivity and occurrence-lift, System P} (1{,}500 observational traces). $n$ is the count of pairs with $d_i > \epsilon$. \emph{Class:} amplifier ($\hat{\sigma} > 1$), absorber ($\hat{\sigma} < 1$). The median ratio is reported alongside the mean to expose multimodal distributions.}
\label{tab:sigma}
\centering
\small
\begin{tabular}{lrrrcr}
\toprule
edge ($v_i \to v_j$) & $n$ & $\hat{\sigma}_{ij}$ & median ratio & class & $\lambda_{ij}$ \\
\midrule
rewriter $\to$ discovery        & 266 & \textbf{1.128} & 0.650 & amplifier & $+0.994$ \\
rewriter $\to$ planner          & 421 & 0.297 & 0.141 & absorber  & $+0.27$  \\
signal\_analysis $\to$ planner  & 109 & 0.857 & 0.327 & absorber  & $+0.041$ \\
discovery $\to$ planner         & 267 & 0.295 & 0.230 & absorber  & $+0.27$  \\
discovery $\to$ composer        & 267 & 0.313 & 0.088 & absorber  & $+0.04$  \\
planner $\to$ composer          & 943 & \textbf{1.069} & 0.537 & amplifier & $+0.14$  \\
\bottomrule
\end{tabular}
\end{table}

\begin{table}[t]
\caption{\textbf{Edge sensitivity, System Q} (24{,}693 same-input pairs). Edges named per the architecture in Figure~\ref{subfig:sys_q}. $n$ is the count of pairs with $d_i > \epsilon$. \emph{$D_{\mathrm{noise}}$} columns report the within-input noise floor at each node. Tool-execution terminals (insensitive, $\hat{\sigma}=0$ from every upstream) are omitted as they carry no per-edge signal. We omit $\lambda_{ij}$ for System~Q because the heterogeneous group-size pairing (production replay, see \S4) yields highly variable $|d_i \leq \epsilon|$ partition sizes per edge, making per-edge $\hat{\lambda}$ estimates noisy and not directly comparable to System~P's balanced-design $\lambda$ values in Table~\ref{tab:sigma}.}
\label{tab:sigma_q}
\centering
\small
\begin{tabular}{lrrcrr}
\toprule
edge ($v_i \to v_j$) & $n$ & $\hat{\sigma}_{ij}$ & class & $D_{\mathrm{noise}}(v_i)$ & $D_{\mathrm{noise}}(v_j)$ \\
\midrule
Fast LLM $\to$ Company Search & 19{,}525 & \textbf{3.635} & amplifier & 0.180 & 0.281 \\
Fast LLM $\to$ Reranker     & 18{,}009 & \textbf{9.018} & amplifier & 0.180 & 0.538 \\
Fast LLM $\to$ Generator    & 18{,}009 & \textbf{3.039} & amplifier & 0.180 & 0.178 \\
Slow LLM $\to$ Company Search & 13{,}574 & \textbf{3.418} & amplifier & 0.172 & 0.281 \\
Slow LLM $\to$ Reranker     & 18{,}008 & \textbf{5.982} & amplifier & 0.172 & 0.538 \\
Slow LLM $\to$ Generator    & 18{,}009 & \textbf{1.920} & amplifier & 0.172 & 0.178 \\
Company Search $\to$ Reranker & 18{,}009 & \textbf{4.092} & amplifier & 0.281 & 0.538 \\
Company Search $\to$ Generator & 18{,}009 & \textbf{1.505} & amplifier & 0.281 & 0.178 \\
Reranker $\to$ Generator    & 18{,}008 & 0.341          & absorber  & 0.538 & 0.178 \\
\bottomrule
\end{tabular}
\end{table}

\subsection{Path Sensitivity Approximation Holds Empirically}
We computed transitive $\hat{\sigma}(i,j)$ for reachable pairs in System~P and compared against the path-product approximation (Definition~6). The multiplicative approximation agrees within $\pm 4\%$ (rewriter $\to$ composer: empirical 0.366 vs.\ product 0.356; signal\_analysis $\to$ composer: 0.879 vs.\ 0.916), indicating interaction terms are negligible for this pipeline. Transitive measurements revealed risk profiles invisible in the direct-edge view: one source node with no direct edge to the output exhibited transitive $\hat{\sigma} = 0.879$ (nearly unity), while another with a direct amplifier edge to retrieval showed transitive $\hat{\sigma} = 0.366$ to the output, dampened by intervening absorbers.

\subsection{Three-Component Divergence Captures Structural Information}

Table~\ref{tab:divergence} reports the trajectory divergence distribution across both corpora and both systems.

\begin{table}[t]
\caption{\textbf{Trajectory divergence rates.} System P observational from 1{,}500 traces; System P interventional under rewriter variant swap (515 baseline-perturbed pairs). System Q from 24{,}693 same-input pairs (1{,}497 traces). Rates are population aggregates over the indicator $\mathbb{1}[D_\bullet > 0]$.}
\label{tab:divergence}
\centering
\small
\begin{tabular}{lrrr}
\toprule
metric & System P (obs.) & System P (interv., rewriter) & System Q (obs.) \\
\midrule
$D_{\mathrm{iter}} > 0$           &  3.3\% &  9.5\% & 23.7\% \\
$D_{\mathrm{shape}} > 0$          & 12.9\% & 37.5\% & 25.1\% \\
$D_{\mathrm{output}} > 0$ (any)   & 92.4\% & — & — \\
$D_{\mathrm{output}}$-only         & 76.3\% & — & — \\
$D_{\mathrm{struct}} > 0$ (node-set divergence) & 16.2\% & 46.9\% & 20.6\% \\
\bottomrule
\end{tabular}
\end{table}

In System~P, the majority of observational divergences (76.3\%) are pure value drift; 12.9\% exhibit structural divergence; 3.3\% exhibit iteration count divergence. Under interventional perturbation, structural divergence approximately triples (12.9\% $\to$ 37.5\% for $D_{\mathrm{shape}}$), confirming that upstream perturbation causally drives trajectory bifurcation. System~Q shows higher baseline structural divergence (25.1\% $D_{\mathrm{shape}}$) and a non-trivial 23.7\% $D_{\mathrm{iter}}$ that comes from sub-stage call-count variability (e.g., reranker invocations, tool-execution counts) rather than architectural loops, the metric captures any per-node count divergence, not only loop iteration counts. The fact that $D_{\mathrm{struct}}$ (20.6\%) is lower than $D_{\mathrm{shape}}$ (25.1\%) for System~Q reflects that some branch-decision differences ($D_{\mathrm{shape}}{>}0$) are within-path: routing parameters or sub-stage decisions vary while the activated node set is unchanged, so $D_{\mathrm{struct}}$ does not fire.

Critically, different perturbation types on System~P produce nearly identical $D_{\mathrm{shape}}$ rates through mechanistically distinct cascades (Table~\ref{tab:cascade}). Rewriter variant swap and signal turn-type flip both yield ${\sim}37\%$ $D_{\mathrm{shape}}$, but the cascade signatures differ: the rewriter cascade amplifies through retrieval then triggers a categorical flip at the planner ($d_{\mathrm{planner}} \approx 0.01$), while the turn-type cascade amplifies directly through the planner ($d_{\mathrm{planner}} \approx 0.27$) into the composer ($d_{\mathrm{composer}} \approx 0.53$). Web search hint perturbation produces the highest $D_{\mathrm{shape}}$ (75.9\%) by inserting a new node into the graph topology. Context sufficiency override produces 63.4\% $D_{\mathrm{shape}}$ in the effective stratum by \emph{removing} a node. These cascade patterns are distinguishable only through per-node sensitivity decomposition.

\begin{table}[t]
\caption{\textbf{Cascade patterns across interventional perturbations on System P.} Same near-boundary seed set (115 utterances), four upstream perturbation classes, four mechanistically distinct cascades. \emph{Cascade pattern} is the qualitative shape of $d$ along the downstream chain conditional on $D_{\mathrm{shape}} > 0$. CS-override rates are reported on the \emph{effective} stratum (262 pairs where the override changed the natural value).}
\label{tab:cascade}
\centering
\small
\setlength{\tabcolsep}{3pt}
\begin{tabular}{lrrlp{0.32\linewidth}}
\toprule
perturbation & $D_{\mathrm{shape}}$ & $D_{\mathrm{iter}}$ & cascade pattern & mechanism \\
\midrule
rewriter variant swap        & 37.5\%   &  9.5\%   & amplify-then-flip       & retrieval boundary $\to$ planner categorical flip \\
signal turn\_type flip       & 36.5\%   &  9.3\%   & direct planner chain    & field consumed in planner prompt directly \\
signal web\_search=certain   & \textbf{75.9\%} & 10.4\% & node insertion (DAG)    & gates web\_search node activation \\
signal is\_small\_talk=true  & 0\%      & \textbf{100\%} & full path collapse  & SHORT\_CIRCUIT branch always taken \\
context\_sufficiency=True (eff.) & \textbf{63.4\%} & 36.6\% & node removal (DAG) & SHORTCUT skips discovery \\
\bottomrule
\end{tabular}
\end{table}

\subsection{Bifurcation Thresholds Are Measurable}

We estimated bifurcation thresholds (Definition~10) under both observational and interventional conditions for System~P.

Under observational estimation, the planner is the sole bifurcation source ($\hat{\beta}_{\mathrm{shape}} = 0.102$); all other nodes show $\hat{\beta} = 0$, meaning structural divergence occurs even when those nodes' outputs are identical across traces. Under interventional perturbation, the rewriter exhibits $\hat{\beta}_{\mathrm{shape}} \approx 0.25$ (upper bound under sparse coverage below 0.4; see \S\ref{sec:limitations}): below this threshold, no structural divergence is observed; above it, 37.5\% of traces bifurcate.

The two estimates are complementary: observational $\beta$ identifies the proximate intrinsic source (planner) for monitoring; interventional $\beta$ identifies the externally-induced threshold (rewriter) for change management. At the boundary, the cascade exhibits amplify-then-flip dynamics: moderate upstream drift ($d_{\mathrm{rewriter}}{\approx}0.225$) amplifies at retrieval ($d_{\mathrm{discovery}}{\approx}0.367$), collapses to a near-zero planner output change ($d_{\mathrm{planner}}{\approx}0.01$) with a categorical field flip, and triggers structural divergence.

\subsection{Cross-Architecture Validation: DSPy Multi-Hop QA}
\label{sec:dspy}

To validate generalization, we instrument a DSPy~\citep{khattab2024dspy} multi-hop QA pipeline on HotpotQA~\citep{yang2018hotpotqa}, a strict sequential chain (\textsc{gen\_query\_1}$\to$\textsc{retrieve\_1}$\to$\textsc{gen\_query\_2}$\to$\textsc{retrieve\_2}$\to$\textsc{gen\_answer}) with no parallel intake, no loops, no conditional branching; 1{,}500 traces (500 dev questions $\times$ 3 repeats), same determinism contract, ColBERTv2~\citep{santhanam2022colbertv2} retrieval over per-question distractor passages (canonical DSPy setup).

Across all 1{,}500 same-input pairs, $D_{\mathrm{iter}}=D_{\mathrm{shape}}=D_{\mathrm{struct}}=0$ --- as predicted for a chain without conditional branches, and a clean separation from Systems~P (12.9\% $D_{\mathrm{shape}}$) and~Q (25.1\%). Value drift remains non-zero ($D_{\mathrm{noise}}$ system mean 0.255), confining variation to the $D_{\mathrm{output}}$-only regime of Definition~9. Edge sensitivities (Table~\ref{tab:sigma_dspy}) split cleanly: the upstream rewriter \textsc{gen\_query\_1} amplifies downstream ($\hat{\sigma}{\approx}3.2$, bimodal); final \textsc{retrieve}$\to$\textsc{gen\_answer} edges absorb sharply ($\hat{\sigma}{\approx}0.28$--$0.31$). A matched-protocol BM25 replication (Appendix~\ref{app:dspy_full}) preserves the sign of dominant amplifiers and absorbers; three near-unity edges ($|\hat{\sigma}-1|<0.4$) cross $\hat{\sigma}{=}1$ between retrievers --- exactly the regime where Definition~3's binary classification is most kernel-sensitive --- yet the framework's decomposition still isolates the mechanism (e.g., \textsc{gen\_query\_1}$\to$\textsc{retrieve\_1}: intrinsic-noise term $0.000$ under both retrievers; median ratio $d_j/d_i$ shifts from $0.26$ to $0.86$, indicating upstream-coupling sensitivity rather than per-node noise; full numbers in Appendix~\ref{app:dspy_full}). Two production interventions (Appendix~\ref{app:case_studies}) reinforce this: framework outputs surfaced pipeline-level couplings invisible to component-level evaluation, driving prompt-level and architectural changes that improved quality and reduced variance.

\begin{table}[t]
\caption{\textbf{Edge sensitivity, DSPy multi-hop QA} (1{,}500 same-input pairs, 500 HotpotQA dev questions $\times$ 3 repeats; ColBERTv2 retrieval). Selected edges; full 10-edge matrix in Appendix~\ref{app:dspy_full}.}
\label{tab:sigma_dspy}
\centering
\small
\begin{tabular}{lrrcrr}
\toprule
edge ($v_i \to v_j$) & $n$ & $\hat{\sigma}_{ij}$ & class & $D_{\mathrm{noise}}(v_i)$ & $D_{\mathrm{noise}}(v_j)$ \\
\midrule
generate\_query\_1 $\to$ generate\_query\_2  & 1{,}344 & \textbf{3.230} & amplifier (bimodal) & 0.233 & 0.312 \\
generate\_query\_1 $\to$ retrieve\_1         & 1{,}344 & \textbf{1.369} & amplifier           & 0.233 & 0.267 \\
retrieve\_1        $\to$ retrieve\_2         &   806   & 0.928          & absorber            & 0.267 & 0.345 \\
retrieve\_1        $\to$ generate\_query\_2  &   808   & 0.823          & absorber            & 0.267 & 0.312 \\
retrieve\_2        $\to$ generate\_answer    &   874   & \textbf{0.282} & strong absorber     & 0.345 & 0.116 \\
\bottomrule
\end{tabular}
\end{table}

\subsection{Distribution Faithfulness Quantifies Per-Node Evaluation Gap}

We computed distribution faithfulness (Principle~1) on 1{,}500 System~P traces, comparing per-node golden vs.\ actual outputs.

\begin{table}[t]
\caption{\textbf{Distribution faithfulness gap per node, System P.} Per-field $d$ between actual run outputs and golden expected outputs averaged across 1{,}500 traces ($\times$ field count). Type-dispatched kernels per Section~2.2; fragment-reference fields use recall-based distance to handle the broad-recall vs.\ relevant-subset asymmetry. \emph{Min/max field} report the per-field range within the node. $^{\dagger}$Discovery's gap of 0.900 reflects a structural retrieval/golden-set asymmetry (broad retrieval set vs.\ smaller golden-relevant subset), not model misalignment.}
\label{tab:faithful}
\centering
\small
\begin{tabular}{lrrrr}
\toprule
node & $n$ & mean gap & min field & max field \\
\midrule
signal\_analysis & 9{,}000 & \textbf{0.047} & 0.000 & 0.225 \\
planner          & 8{,}856 & 0.347 & 0.034 & 0.932 \\
rewriter         & 1{,}500 & 0.646 & 0.646 & 0.646 \\
composer         & 4{,}356 & 0.696 & 0.410 & 0.959 \\
discovery$^{\dagger}$ & 1{,}356 & 0.900 & 0.900 & 0.900 \\
\midrule
\textbf{system} & — & \textbf{0.527} & — & — \\
\bottomrule
\end{tabular}
\end{table}

Gaps span an order of magnitude: 0.047 (signal analysis, well-characterized) to 0.696 (composer); discovery's recall-based gap of 0.900 reflects retrieval/golden-set asymmetry rather than model misalignment. Per-field variation within nodes is also substantial (planner routing fields 0.03--0.16; context fields 0.48--0.93).

%% file: limitations.tex
\section{Conclusion, Limitations, and Future Work}
\label{sec:limitations}
QUIVER provides primitives: $\sigma$, $\lambda$, three-component divergence, $\beta$, distribution faithfulness,  that capture propagation dynamics invisible to component-level and end-to-end evaluation; validation across Systems~P, Q, and DSPy shows architecture-appropriate measurements with no modification. \emph{Limitations:} $\hat{\sigma}$ is locally valid; re-estimation is needed after large prompt/model/index changes. Rewriter drift covered $[0.696, 0.754]$; $\hat{\beta}_{\mathrm{shape}} \approx 0.25$ is an upper bound under sparse coverage below $0.4$. Definitions~2--4 are kernel-/weight-agnostic; reported values are one instantiation. \emph{Future work:} data-driven weight estimation, broader pipeline classes, non-conversational domains; broader impact in Appendix~\ref{app:broader}.

%% file: appendix1.tex
\appendix

\section{Extended Discussion of Definitions}
\label{app:definitions}

This appendix provides extended discussion, interpretation, and worked examples for each formal definition in Section~2.

\subsection{Typed Pipeline Graph (Definition 1)}

The typed pipeline graph $\mathcal{G} = (V, E, \mathcal{T}, \mathcal{F})$ captures the essential structure of any compound LLM pipeline. Each node $v_i$ represents a computational unit --- typically an LLM call, a retrieval operation, a deterministic transformation, or a routing decision. The stochastic function $f_i : \prod_{j \in \mathrm{pa}(i)} T_j \rightarrow \Delta(T_i)$ maps parent outputs to a \emph{distribution} over outputs rather than a single output, reflecting the inherent non-determinism of LLM calls (temperature, sampling, backend non-determinism) and stochastic retrieval operations.

The typed output spaces $T_i$ are determined by each node's implementation:
\begin{itemize}[nosep]
    \item \textbf{Schema-typed objects:} JSON conforming to a defined schema (e.g., a planner output with fields for task, intent type, remaining subtasks, and reasoning trace). Formally, $T_i = T_i^{(1)} \times \cdots \times T_i^{(m)}$ is a product of $m$ typed fields.
    \item \textbf{Ordered lists:} Ranked sequences with typed elements (e.g., a retrieval node returning a ranked list of document fragments).
    \item \textbf{Categorical values:} Elements from a finite set (e.g., an intent classifier outputting one of $\{\texttt{FAQ}, \texttt{troubleshoot}, \texttt{request}\}$).
    \item \textbf{Unstructured text:} Free-form natural language (e.g., a composer node generating a response, or a rewriter node reformulating a query).
\end{itemize}

A single pipeline may combine all of these types across different nodes. External inputs to the pipeline (user queries, session state, conversation history, configuration parameters) are modeled as source nodes $v_s$ with $\mathrm{pa}(s) = \emptyset$, whose output distributions $\Delta(T_s)$ are determined by the deployment environment rather than by a learned function.

\paragraph{Worked example.} Consider a pipeline with five nodes: Rewriter ($v_1$), Signal Analysis ($v_2$), Discovery ($v_3$), Planner ($v_4$), Composer ($v_5$). The typed output spaces are:
\begin{align*}
T_1 &= \texttt{RewriterOutput} = \{\texttt{intents}: \texttt{Map[str, List[str]]}\} \\
T_2 &= \texttt{SignalResult} = \{\texttt{turn\_type}: \texttt{Cat}, \; \texttt{sentiment}: \texttt{Cat}, \; \texttt{web\_hint}: \texttt{Cat}, \; \texttt{is\_small\_talk}: \texttt{Bool}\} \\
T_3 &= \texttt{DiscoveryOutput} = \{\texttt{fragment\_ids}: \texttt{List[str]}\} \\
T_4 &= \texttt{PlannerOutput} = \{\texttt{intent\_type}: \texttt{Cat}, \; \texttt{task}: \texttt{str}, \; \texttt{remaining}: \texttt{List[str]}, \; \texttt{gaps}: \texttt{List[str]}, \; \texttt{thought}: \texttt{str}\} \\
T_5 &= \texttt{ComposerOutput} = \{\texttt{response}: \texttt{str}, \; \texttt{selected\_refs}: \texttt{List[str]}\}
\end{align*}

The edge set $E$ encodes which nodes consume which outputs: $E = \{(v_1, v_3), (v_1, v_4), (v_2, v_4), (v_3, v_4), (v_3, v_5), (v_4, v_5)\}$, indicating that the Planner ($v_4$) has four parents and the Composer ($v_5$) has two.

\subsection{Type-Dispatched Distance Metrics (Definition 2)}

The key design principle is that distance metrics are \emph{dispatched by type}: different output types require fundamentally different notions of distance. Embedding cosine distance between two JSON objects with one categorical field flipped tells you almost nothing; a type-aware metric that checks ``did the enum value change?'' gives a precise binary signal.

For schema-typed output spaces, the distance decomposes as a weighted sum of per-field distances:
\begin{equation}
d_i(x, y) = \sum_{k=1}^{m} w_k \cdot d_i^{(k)}(x^{(k)}, y^{(k)})
\end{equation}

The per-field distance functions are:
\begin{itemize}[nosep]
    \item \textbf{Categorical fields} (e.g., intent classification, action type): discrete metric $d(a,b) = \mathbb{1}[a \neq b]$. A classification flip from \texttt{FAQ} to \texttt{troubleshoot} has distance 1; same classification has distance 0.
    \item \textbf{Boolean fields} (e.g., \texttt{is\_small\_talk}): discrete metric $d(a,b) = \mathbb{1}[a \neq b]$.
    \item \textbf{Set-valued fields} (e.g., list of fragment IDs, set of retrieved document identifiers): normalized Jaccard distance $d(A,B) = 1 - |A \cap B| / |A \cup B|$. Two fragment sets sharing 15 of 20 elements have distance $1 - 15/25 = 0.4$.
    \item \textbf{Ordered list fields} (e.g., ranked results, remaining task list): normalized edit distance or rank correlation distance, depending on whether order carries semantic meaning.
    \item \textbf{Numeric fields} (e.g., confidence scores, response length): normalized absolute difference $d(a,b) = |a - b| / \max(|a|, |b|, \epsilon)$.
    \item \textbf{Text fields} (e.g., reasoning traces, generated responses): embedding cosine distance $d(s,t) = 1 - \cos(\phi(s), \phi(t))$ for an embedding function $\phi$. We use \texttt{all-MiniLM-L6-v2} throughout our experiments.
    \item \textbf{Mapping fields} (e.g., \texttt{Map[category, List[query]]}): combined key-set Jaccard plus per-key text distance, averaged.
\end{itemize}

\paragraph{Field weights.} The weights $w_k$ reflect each field's causal influence on downstream nodes rather than its semantic importance to a human reader. We distinguish three categories:
\begin{itemize}[nosep]
    \item \textbf{Routing fields:} Fields consumed by downstream nodes in programmatic routing logic or conditional branching (e.g., \texttt{intent\_type} determining which agent executes, \texttt{remaining} determining loop continuation). Assigned weight $2w_0$.
    \item \textbf{Context fields:} Fields consumed by downstream nodes as prompt context without affecting control flow (e.g., \texttt{task} description, \texttt{fragment\_refs}). Assigned weight $w_0$.
    \item \textbf{Observability fields:} Fields used only for logging, debugging, or human inspection, not consumed by any downstream node (e.g., \texttt{thought} reasoning traces). Assigned weight $0$.
\end{itemize}
where $w_0$ is a normalizing constant ensuring $\sum_k w_k = 1$. This assignment can be determined by static analysis of the pipeline code: trace which fields appear in downstream prompt templates or routing conditions. Specific weight ratios are application-dependent; the framework's measurements are defined for any admissible choice.

\subsection{Edge Sensitivity (Definition 3)}

The edge sensitivity coefficient $\sigma_{ij}$ quantifies the expected ratio of downstream output distance to upstream output distance:
\begin{equation}
\sigma_{ij} = \mathbb{E}\left[\frac{d_j\big(f_j(\mathbf{x}),\; f_j(\mathbf{x}')\big)}{d_i(x_i, x_i')}\right]
\end{equation}

The expectation is taken over the distribution of all other parent inputs and the stochasticity of $f_j$, restricted to pairs where $d_i(x_i, x_i') > \epsilon$ for a small threshold $\epsilon > 0$ to avoid degenerate ratios when the upstream outputs are nearly identical.

\paragraph{Classification.} Edge sensitivity classifies each connection in the pipeline graph:
\begin{itemize}[nosep]
    \item $\sigma_{ij} > 1$: edge $(v_i, v_j)$ is an \textbf{amplifier} --- node $v_j$ magnifies upstream variation from $v_i$.
    \item $\sigma_{ij} < 1$: edge $(v_i, v_j)$ is an \textbf{absorber} --- node $v_j$ dampens upstream variation from $v_i$.
    \item $\sigma_{ij} \approx 0$: node $v_j$ is \textbf{insensitive} to $v_i$ --- the edge carries data but does not influence behavior.
\end{itemize}

\paragraph{Threshold-sensitive edges.} The scalar $\sigma_{ij}$ is a summary statistic that may obscure richer structure. Our empirical results (Section~5) reveal that edges classified as ``amplifiers'' by their mean $\hat{\sigma}$ can exhibit bimodal ratio distributions: a majority regime where the edge absorbs (ratio $< 1$) and a minority tail regime where the edge amplifies by factors of $3$--$7\times$. The mean lands near unity because the two regimes roughly balance. This bimodality reflects \emph{threshold-sensitive coupling}: below a domain-specific boundary (e.g., an embedding similarity threshold in a retrieval index), the edge absorbs perturbations; above it, the edge amplifies. We recommend inspecting the full ratio distribution alongside the scalar summary to identify threshold-sensitive edges.

\subsection{Occurrence-Lift (Definition 4)}

Occurrence-lift $\lambda_{ij}$ captures a complementary aspect of perturbation propagation that $\sigma_{ij}$ misses: the \emph{probability} that downstream drift occurs at all, independent of its magnitude.

\begin{equation}
\lambda_{ij} = P\big(d_j > 0 \mid d_i > 0\big) - P\big(d_j > 0 \mid d_i = 0\big)
\end{equation}

\noindent The thresholds are written as $d > 0$ for clarity; under continuous-valued LLM outputs the estimator partitions on a small $\epsilon > 0$ matched to the per-type kernel's numerical floor (Appendix~\ref{app:estimation}, B.2), and we treat $d > \epsilon$ as the operative event throughout.

The two measures are decoupled, carrying distinct information about an edge:
\begin{itemize}[nosep]
    \item \textbf{High $\sigma$, low $\lambda$:} When upstream drift does propagate, it propagates strongly, but it rarely does. Example: a signal analysis node with very low intrinsic variance ($D_{\mathrm{noise}} = 0.008$) but high sensitivity to planner when it does drift ($\sigma = 0.86$).
    \item \textbf{Low $\sigma$, high $\lambda$:} Upstream drift almost always triggers some downstream drift, but the magnitude is small. Example: a rewriter-to-discovery edge where any rewrite change affects which fragments are retrieved ($\lambda = +0.994$) but the number of changed fragments is moderate ($\sigma = 1.13$).
    \item \textbf{High $\sigma$, high $\lambda$:} Strong coupling in both probability and magnitude --- the most dangerous edges.
    \item \textbf{Low $\sigma$, low $\lambda$:} Effectively decoupled --- the edge carries data but does not influence downstream behavior.
\end{itemize}

A framework measuring only $\sigma$ (magnitude) or only $\lambda$ (probability) provides an incomplete picture of edge behavior. The sensitivity matrix $\Sigma$ and the occurrence-lift matrix $\Lambda$ together characterize the full propagation dynamics.

\subsection{Path Sensitivity (Definition 6)}

The path sensitivity $\sigma(p) = \prod \sigma_{i_l, i_{l+1}}$ along a directed path follows from the chain rule: if node A's output changes by 1 unit and the A$\to$B edge has $\sigma = 2$, and the B$\to$C edge has $\sigma = 0.5$, then the net effect from A to C is $2 \times 0.5 = 1$.

The product formulation makes two assumptions:
\begin{enumerate}[nosep]
    \item \textbf{Local linearity:} The sensitivity of each edge is independent of perturbation magnitude. This is valid for small perturbations around the current operating point. For large perturbations, the bifurcation analysis (Definitions 9--10) provides the appropriate tool.
    \item \textbf{Path independence:} The path-product gives per-path sensitivity, not total sensitivity when multiple paths connect two nodes. For node pairs connected by $k$ paths, we report both $\max_p \sigma(p)$ and the empirically measured transitive sensitivity $\hat{\sigma}(i,j)$. Agreement between the two validates the multiplicative approximation; divergence indicates significant interaction effects.
\end{enumerate}

\paragraph{Parallel reconvergence.} When multiple parallel nodes feed a common downstream node, their joint effect depends on whether perturbations are independent or correlated. The full model includes interaction terms:
\begin{equation}
d_j = \sum_{i \in \mathrm{pa}(j)} \alpha_i\, d_i + \sum_{i < k \in \mathrm{pa}(j)} \gamma_{ik}\, d_i\, d_k + \varepsilon
\end{equation}
where $\alpha_i$ are partial sensitivity contributions, $\gamma_{ik}$ capture pairwise interactions, and $\varepsilon$ captures intrinsic stochasticity and unmeasured influences. Under independence, the joint sensitivity is approximately:
\begin{equation}
\sigma_{\mathrm{joint}}(v_j) = \sqrt{\sum_{i \in \mathrm{pa}(j)} \sigma_{ij}^2}
\end{equation}

If the interaction coefficients $\gamma_{ik}$ are significant, co-occurring perturbations have reinforcing (positive $\gamma$) or compensating (negative $\gamma$) effects. In our experiments, the path-product approximation matched empirical transitive sensitivity within $\pm 4\%$, suggesting interaction terms are negligible along sequential paths in the pipelines studied.

\noindent The closed-form $\sigma_{\mathrm{joint}}(v_j) = \sqrt{\sum_i \sigma_{ij}^2}$ above is the \emph{independence baseline}: it holds when the parents' perturbations are uncorrelated. In practice, parallel parents downstream of a shared upstream node (as in System~P, where multiple intake nodes are conditioned on the same user query) have correlated perturbations, and the independence form is an approximation. The $\pm 4\%$ path-product validation covers \emph{sequential} multiplicativity, not reconvergent independence; we therefore flag the closed-form $\sigma_{\mathrm{joint}}$ as a limitation and recommend the full multivariate regression with interaction terms $\gamma_{ik}$ for reconvergent nodes, treating the closed-form as a reference value to compare against the empirical transitive sensitivity $\hat{\sigma}(i, j)$.

\subsection{Loop-Augmented Pipelines (Definitions 7--8)}

The loop body $L \subseteq V$ contains the nodes that execute at each iteration. At each iteration $t$, the loop controller selects an action $a^{(t)} \in \mathcal{A}$ that determines which subgraph is realized. The action set $\mathcal{A}$ is determined by the pipeline implementation:
\begin{itemize}[nosep]
    \item \texttt{EXECUTE}: invoke a tool or external agent and incorporate the result.
    \item \texttt{RETRY}: re-invoke discovery or retrieval nodes with targeted, gap-filling queries.
    \item \texttt{GENERATE}: produce intermediate artifacts (summaries, sub-answers).
    \item \texttt{COMPOSE}: synthesize a response from accumulated context.
\end{itemize}

Different actions realize different subgraphs within the iteration. A \texttt{RETRY} action adds retrieval nodes to the iteration's realized graph; an \texttt{EXECUTE} action adds tool execution nodes; a \texttt{COMPOSE} action routes directly to the response composer. The iteration topology $g^{(t)} = (a^{(t)}, Q^{(t)})$ captures both the action and its parameters (e.g., which queries were used for retry, which tool was invoked).

The loop terminates when a termination condition is met (e.g., the remaining task list is empty) or the maximum iteration count $k_{\max}$ is reached. The realized iteration count $k^*$ is itself a random variable that depends on all upstream outputs.

\subsection{Trajectory Divergence (Definition 9)}

The three-component trajectory divergence $D(\tau, \tau') = (D_{\mathrm{iter}}, D_{\mathrm{shape}}, D_{\mathrm{output}})$ captures qualitatively different kinds of pipeline variation:

\paragraph{Value divergence ($D_{\mathrm{output}} > 0$, $D_{\mathrm{shape}} = D_{\mathrm{iter}} = 0$).} The pipeline took the same structural path but produced different output values. This is the most common form of variation and the only kind visible to end-to-end evaluation. Causes include LLM sampling variation, retrieval index updates, and upstream drift within the absorber regime of threshold-sensitive edges.

\paragraph{Structural divergence ($D_{\mathrm{shape}} > 0$).} The pipeline executed different internal logic at one or more iterations --- different actions chosen, different nodes activated, different tools invoked. The end-to-end output may be similar or dramatically different, but the \emph{path} was different. This is invisible to evaluation methods that examine only inputs and outputs.

\paragraph{Extent divergence ($D_{\mathrm{iter}} > 0$).} At least one node was invoked a different number of times across the two traces ($\sum_i |c_i - c_i'| > 0$). The motivating case is loop-iteration divergence --- the loop body $L$ runs $k^*$ times in one trace and $k^{*'}$ in the other, so every node $v_i \in L$ contributes $|k^* - k^{*'}|$ to the sum and entire subgraphs exist in one trace but not the other. The same indicator also fires in non-loop pipelines whenever sub-stage call counts differ (e.g., a reranker invoked twice vs.\ three times), without any subgraph being absent. This is the strongest form of structural divergence in the loop case and a per-node call-count divergence indicator in general.

\paragraph{Applicability to non-loop pipelines.} $D_{\mathrm{iter}}$ counts per-node call-count divergence, not only loop iteration counts; for pipelines without explicit loops it is zero whenever each node is invoked the same number of times across $\tau, \tau'$, but takes positive values when sub-stage call counts (e.g., reranker invocations, tool-execution counts) vary --- as observed for System~Q (23.7\%) despite the absence of an architectural loop. $D_{\mathrm{shape}}$ similarly remains meaningful in the loop-free case: it captures differences in conditional branching decisions (e.g., whether a node was activated or skipped, whether a fast or slow path was taken); System~Q exhibits 25.1\% $D_{\mathrm{shape}}$ from such conditional routing.

\subsection{Bifurcation Threshold (Definition 10)}

Bifurcation thresholds identify the perturbation magnitudes at which the pipeline's behavior transitions from continuous degradation to structural path change.

$\beta_{\mathrm{shape}}(v_i)$ is the smallest perturbation at node $v_i$ that causes any downstream structural divergence (different node activations, different branching decisions). $\beta_{\mathrm{iter}}(v_i)$ is the smallest perturbation that causes a different number of loop iterations.

\paragraph{Observational vs.\ interventional estimation.} These thresholds can be estimated in two complementary modes:
\begin{itemize}[nosep]
    \item \textbf{Observational:} From production traces under natural operating conditions. Identifies the proximate source of structural instability --- which node's intrinsic variance is responsible for the observed structural divergence. In our experiments, observational estimation identified the planner as the sole bifurcation source ($\hat{\beta}_{\mathrm{shape}} = 0.102$) while all other nodes showed $\hat{\beta} = 0$ (structural divergence occurred even with zero drift at those nodes).
    \item \textbf{Interventional:} From controlled perturbation experiments. Identifies the causal threshold for externally-induced bifurcation. In our experiments, interventional estimation revealed a rewriter bifurcation threshold of $\hat{\beta}_{\mathrm{shape}} \approx 0.25$ --- dormant under natural conditions but activated when perturbation exceeds this magnitude.
\end{itemize}

The two modes answer different questions. Observational $\beta$ answers: ``Under current operating conditions, which node is causing structural instability?'' Interventional $\beta$ answers: ``If I change this node's prompt, how large a change can I make before the pipeline takes a structurally different path?'' Practitioners need both: observational for monitoring, interventional for change management.

\subsection{Noise Origins}

A node $v_i$ is a \emph{noise origin} if it produces different outputs on identical inputs:
\begin{equation}
P(d_i > \epsilon \mid d_j = 0\;\forall\, j \in \mathrm{pa}(i)) > 0
\end{equation}

Otherwise it is a \emph{propagator} --- its output varies only because its inputs varied. This partitioning is determined by examining trace pairs where all upstream outputs are byte-identical ($d_j \leq \epsilon$ for all parents $j$) and checking whether the node's own output differs.

Noise origins are the root causes of system-level non-determinism. Even with temperature${}=0$ and fixed random seeds, LLM API calls may exhibit backend non-determinism (batching, hardware floating-point variations, load balancing across replicas). The noise origin analysis distinguishes this intrinsic variance from propagated variance inherited from upstream.

This provides a complementary diagnostic to sensitivity analysis. The sensitivity matrix tells you how perturbations propagate; the noise origin analysis tells you where perturbations originate. Together, they provide a complete picture: noise originates at origin nodes, propagates through the graph along high-$\sigma$ edges, and potentially triggers structural bifurcation at nodes with low $\beta_{\mathrm{shape}}$.

This analysis is conceptually related to UProp's~\citep{duan2025uprop} decomposition of uncertainty into intrinsic and extrinsic components, but achieved through a simpler, more practical measurement: a binary partition based on upstream cleanliness, rather than mutual information estimation requiring token-level logprobs.

\subsection{Distribution Faithfulness (Principle 1)}

Distribution faithfulness formalizes a necessary condition for valid per-node evaluation. Let $P_{\mathrm{prod}}(T_i)$ be the marginal distribution over node $v_i$'s output space induced by running the full pipeline on production inputs. Let $P_{\mathrm{eval}}(T_i)$ be the distribution of inputs used when evaluating any downstream node $v_j$ where $(v_i, v_j) \in E$.

A per-node evaluation is \emph{distribution-faithful} with respect to edge $(v_i, v_j)$ if:
\begin{equation}
D_{\mathrm{KL}}(P_{\mathrm{prod}}(T_i) \,\|\, P_{\mathrm{eval}}(T_i)) < \delta
\end{equation}

The practical failure mode: if a golden evaluation dataset assumes clean, well-formed upstream outputs but the upstream node actually produces noisy, partially correct outputs 30\% of the time, the evaluation overestimates the downstream node's quality by testing it on inputs it never encounters in production.

The sensitivity matrix determines where distribution gaps matter most:
\begin{itemize}[nosep]
    \item A faithfulness gap $\Delta$ at a high-sensitivity edge ($\sigma_{ij} \gg 1$) produces a correspondingly amplified quality estimation error at downstream nodes.
    \item The same gap $\Delta$ at a low-sensitivity edge ($\sigma_{ij} \ll 1$) has minimal downstream impact.
    \item A uniform re-evaluation policy would prioritize nodes by gap magnitude alone; the sensitivity-weighted analysis correctly identifies where gaps matter most.
\end{itemize}

Distribution faithfulness is measurable per-field: the gap may vary dramatically across fields of the same node's output. Routing fields may be well-characterized by golden data while context fields diverge substantially, or vice versa. The per-field breakdown, combined with the field weight assignments from Definition~2, determines the effective faithfulness gap.

\subsection{Cross-Stage Regression Detection (Principle 2)}

The impact set $\mathcal{I}(v_i, \alpha) = \{v_j : \sigma(p) > \alpha \text{ for any directed path } p \text{ from } v_i \text{ to } v_j\}$ converts the sensitivity matrix into a regression testing policy.

\paragraph{Operational protocol.} When the prompt or configuration of node $v_i$ is modified:
\begin{enumerate}[nosep]
    \item Measure $d_i$ on a held-out probe set before and after the change.
    \item For each downstream edge $(v_i, v_j)$, compare $d_i$ against the drift budget $\tau_{ij}$ at the team's chosen significance level $\alpha$.
    \item If $d_i > \tau_{ij}$, re-run golden dataset evaluation for node $v_j$.
    \item If $d_i \leq \tau_{ij}$, skip re-evaluation of $v_j$ --- the perturbation is within the safe budget.
\end{enumerate}

This replaces the current industry practice of either re-evaluating everything (expensive and slow) or re-evaluating nothing downstream (risky) with a quantitatively grounded policy derived from the pipeline's measured sensitivity profile.

For pipelines with conditional loops, the impact set should additionally include any loop-body node where $\beta_{\mathrm{shape}}(v_i)$ is below the measured perturbation magnitude $d_i$. These are nodes where the change may trigger structural trajectory divergence, not merely continuous quality degradation.

\paragraph{Drift budgets.} The per-edge drift budget $\tau_{ij}$ is defined as the smallest upstream drift above which the downstream node is likely to exceed its own noise floor:
\begin{equation}
\tau_{ij}@\alpha = \min\{d_i : P(d_j > D_{\mathrm{noise}}(v_j) \mid d_i > \tau) \geq \alpha\}
\end{equation}

Higher $\tau$ indicates a more robust edge. Drift budgets exhibited a wide range across edges in our experiments: from zero (any upstream drift at all triggers downstream risk) to effectively infinite (the downstream node is insensitive at any realistic drift magnitude). This variation confirms that a uniform regression-testing policy is suboptimal.

\section{Estimation Procedure Details}
\label{app:estimation}

\subsection{Observational Estimation of $\sigma_{ij}$}

Given a corpus of $N$ production traces with logged intermediate outputs at every node, we estimate edge sensitivity via the following procedure:

\begin{enumerate}[nosep]
    \item \textbf{Pair formation.} Group traces by input similarity (same seed, same intent class, or similar query embedding). Within each group, form all pairwise combinations. For a corpus of $N$ traces with $S$ seeds and $R$ repeats per seed, this yields $S \binom{R}{2}$ pairs.
    \item \textbf{Distance computation.} For each pair and each node $v_i$, compute the type-dispatched distance $d_i$ using Definition~2. Store the per-field distances alongside the aggregate for later analysis.
    \item \textbf{Sensitivity estimation.} For each edge $(v_i, v_j) \in E$, collect all pairs where $d_i > \epsilon$ and compute $\hat{\sigma}_{ij} = \mathbb{E}[d_j / d_i]$ over those pairs.
    \item \textbf{Partial regression (for multi-parent nodes).} When node $v_j$ has multiple parents, the simple ratio $d_j / d_i$ confounds the contributions of different parents. We estimate partial sensitivities via multivariate regression:
    \begin{equation}
    d_j = \sum_{k \in \mathrm{pa}(j)} \alpha_k\, d_k + \sum_{k < l \in \mathrm{pa}(j)} \gamma_{kl}\, d_k\, d_l + \varepsilon
    \end{equation}
    The coefficient $\alpha_i$ is the partial sensitivity of $v_j$ to $v_i$, controlling for co-variation across other parents. The interaction terms $\gamma_{kl}$ capture pairwise reinforcing or compensating effects.
\end{enumerate}

\paragraph{Sample size.} The partial regression requires $N$ sufficiently large relative to $|\mathrm{pa}(j)|^2$ (the number of parameters including interaction terms). For nodes with 5 parents, this is 15 parameters (5 main effects + 10 interactions), requiring on the order of hundreds of pairs for stable estimates. In our experiments, 1{,}500 traces (yielding 266--943 qualifying pairs per edge depending on $\epsilon$ filtering) provided stable estimates for System~P with 6 edges and a maximum of 4 parents per node.

\paragraph{Epsilon threshold.} The threshold $\epsilon$ avoids degenerate ratios when upstream outputs are nearly identical. We use a fixed $\epsilon = 0.01$ throughout. When the upstream node's noise floor $D_{\mathrm{noise}}(v_i)$ is very small (e.g., $< 0.01$), few pairs qualify and estimates may have wide confidence intervals. We report sample sizes alongside all sensitivity estimates.

\subsection{Observational Estimation of $\lambda_{ij}$, $\tau_{ij}$, and $D_{\mathrm{noise}}$}

\paragraph{Occurrence-lift.} For each edge, partition pairs into those with $d_i > \epsilon$ and those with $d_i \leq \epsilon$ (using the same per-type kernel floor as in $\hat{\sigma}_{ij}$). Compute the conditional probabilities $P(d_j > \epsilon \mid d_i > \epsilon)$ and $P(d_j > \epsilon \mid d_i \leq \epsilon)$; the difference is $\hat{\lambda}_{ij}$.

\paragraph{Drift budgets.} For each edge and each significance level $\alpha$, sweep upstream drift thresholds $\tau$ and compute $P(d_j > D_{\mathrm{noise}}(v_j) \mid d_i > \tau)$. The smallest $\tau$ at which this probability reaches $\alpha$ is the drift budget $\tau_{ij}@\alpha$. When the probability never reaches $\alpha$ at any observed drift magnitude, the edge is effectively insensitive at that significance level (reported as ``never'' in our tables).

\paragraph{Noise floor.} $D_{\mathrm{noise}}(v_i)$ is the mean type-dispatched distance across same-input pairs: $D_{\mathrm{noise}}(v_i) = \mathbb{E}[d_i(o_i, o_i')]$ over pairs with identical pipeline inputs. This measures the intrinsic non-determinism of each node under the fixed determinism contract (temperature, seed).

\subsection{Interventional Estimation of $\beta_{\mathrm{shape}}$ and $\beta_{\mathrm{iter}}$}

Bifurcation thresholds describe decision boundaries and cannot be reliably estimated from observational data alone, as the natural variation may not probe the boundary region.

\begin{enumerate}[nosep]
    \item \textbf{Boundary identification.} From the observational corpus, identify traces near the bifurcation boundary --- traces where the loop controller nearly chose a different action or a conditional branch was marginal. Heuristics include: gap analysis that returned marginal results, confidence scores near decision thresholds, or categorical classifications with low model confidence.
    \item \textbf{Perturbation application.} For each near-boundary trace and each target upstream node $v_i$, apply controlled perturbations of known magnitude. Perturbation strategies depend on the node's output type:
    \begin{itemize}[nosep]
        \item \textbf{Text outputs:} paraphrase at varying temperatures, keyword addition/removal, prompt variant substitution.
        \item \textbf{Categorical outputs:} flip to next-most-likely class.
        \item \textbf{List outputs:} remove elements, reorder, add spurious elements.
        \item \textbf{Boolean outputs:} flip the value.
    \end{itemize}
    \item \textbf{Re-execution.} Re-run the pipeline from the perturbed node forward, holding all other inputs fixed.
    \item \textbf{Threshold recording.} Record the smallest perturbation magnitude $d_i$ at which $D_{\mathrm{shape}} > 0$ or $D_{\mathrm{iter}} > 0$.
    \item \textbf{Aggregation.} Report $\hat{\beta}_{\mathrm{shape}}(v_i)$ as the minimum observed threshold across all near-boundary traces, with sample size and interquartile range.
\end{enumerate}

\paragraph{Stratification.} For perturbations that are binary (categorical flips, boolean overrides), the perturbation is either effective (the baseline had a different value) or a no-op (the baseline already matched the perturbed value). We report these strata separately: the effective stratum gives the bifurcation rate when the perturbation actually changes something; the no-op stratum serves as a built-in negative control, where all divergence metrics should return to the noise floor.

\paragraph{Limitations.} The precision of $\hat{\beta}_{\mathrm{shape}}$ depends on the resolution of perturbation magnitudes available. Discrete perturbations (categorical flips, prompt variant swaps) provide binary probe points; continuous perturbations (paraphrasing at varying temperatures) provide finer resolution but are harder to control precisely. Our rewriter perturbation experiments used prompt variant swaps, yielding mean $d_{\mathrm{rewriter}} \in [0.696, 0.754]$ with sparse coverage below 0.4. The reported threshold $\hat{\beta}_{\mathrm{shape}} \approx 0.25$ is therefore an upper bound; the true threshold may be lower.

\subsection{Noise Origin Classification}

For each pair $(a, b)$ and each node $v_i$, we partition by whether every recorded upstream output is byte-identical ($d_j \leq \epsilon$ for all $j \in \mathrm{pa}(i)$):
\begin{itemize}[nosep]
    \item If $d_i > \epsilon$ when all upstream outputs are identical, $v_i$ is an \textbf{origin} --- it generates variance intrinsically.
    \item If $d_i > \epsilon$ only when at least one upstream output differs, $v_i$ is a \textbf{propagator} --- it inherits variance.
    \item If $v_i$ never has all-clean upstream (because at least one parent is always dirty), the classification is indeterminate from the available data. We report these as ``always upstream-dirty'' and note the drift rate when upstream is dirty.
\end{itemize}

\subsection{Distribution Faithfulness Estimation}

For each node $v_i$ with a corresponding golden evaluation field, we compute the per-field type-dispatched distance between each production trace output and the golden expected output. The per-node distribution gap is the mean across fields; the per-field gap identifies which specific output dimensions are well-characterized versus poorly-characterized by the golden data.

We use the same type-dispatched kernel family as Definition~2; the specific instantiation (text-distance, list-distance, etc.) is application-dependent and orthogonal to the per-node faithfulness gap.

\section{Capability Comparison with Related Work}
\label{app:capability}

Table~\ref{tab:comparison} compares capability coverage across the most closely related approaches to QUIVER. Symbols: \cmark = supported, \pmark = partial, \xmark = not supported.

\begin{table}[h]
\caption{\textbf{Capability comparison with related work.} QUIVER is the only approach providing structural bifurcation analysis and distribution faithfulness without requiring auxiliary models.}
\label{tab:comparison}
\centering
\small
\setlength{\tabcolsep}{4pt}
\begin{tabular}{lccccccccc}
\toprule
 & \rotatebox{65}{\shortstack{Computation\\graph}} 
 & \rotatebox{65}{\shortstack{Typed\\distances}} 
 & \rotatebox{65}{\shortstack{Continuous\\$\sigma$}} 
 & \rotatebox{65}{\shortstack{Conditional\\loops}} 
 & \rotatebox{65}{\shortstack{Bifurcation\\detection}} 
 & \rotatebox{65}{\shortstack{Type-dispatched\\attribution}} 
 & \rotatebox{65}{\shortstack{Distribution\\faithfulness}} 
 & \rotatebox{65}{\shortstack{No auxiliary\\model}} \\
\midrule
\rowcolor{gray!10} \textbf{QUIVER (ours)} & \cmark & \cmark & \cmark & \cmark & \cmark & \cmark & \cmark & \cmark \\
UProp                    & \xmark & \xmark & \cmark & \xmark & \xmark & \xmark & \xmark & \xmark \\
AgenTracer               & \pmark & \xmark & \xmark & \pmark & \xmark & \xmark & \xmark & \xmark \\
Spark to Fire            & \cmark & \xmark & \cmark & \xmark & \xmark & \xmark & \xmark & \cmark \\
He et al.                & \pmark & \xmark & \cmark & \xmark & \xmark & \xmark & \xmark & \xmark \\
Geometric Dyn.           & \xmark & \xmark & \cmark & \cmark & \pmark & \xmark & \xmark & \cmark \\
DSPy / TextGrad          & \cmark & \pmark & \xmark & \xmark & \xmark & \xmark & \xmark & \cmark \\
\bottomrule
\end{tabular}
\end{table}

\noindent The partial (\pmark) for DSPy/TextGrad under \emph{Typed distances} reflects that DSPy signatures \emph{declare} field types and TextGrad supports textual feedback over typed I/O, but neither dispatches a type-appropriate distance kernel per field (categorical, set-valued, ordered-list, numeric, text) nor aggregates them into a per-edge $\sigma$; the type information is used for prompt scaffolding and parsing rather than for measurement.

\section{Broader Impact}
\label{app:broader}

The framework provides diagnostic capability for production AI systems, enabling practitioners to identify and remediate fragile components before they cause user-facing failures. The sensitivity matrix could theoretically be used to identify vulnerable edges for adversarial exploitation, but this risk is minimal since the measurements require access to internal pipeline traces that external adversaries would not have.

\section{DSPy Multi-Hop QA --- Cross-Retriever Sensitivity}
\label{app:dspy_full}

\begin{figure}[h]
\centering
\includegraphics[width=\textwidth]{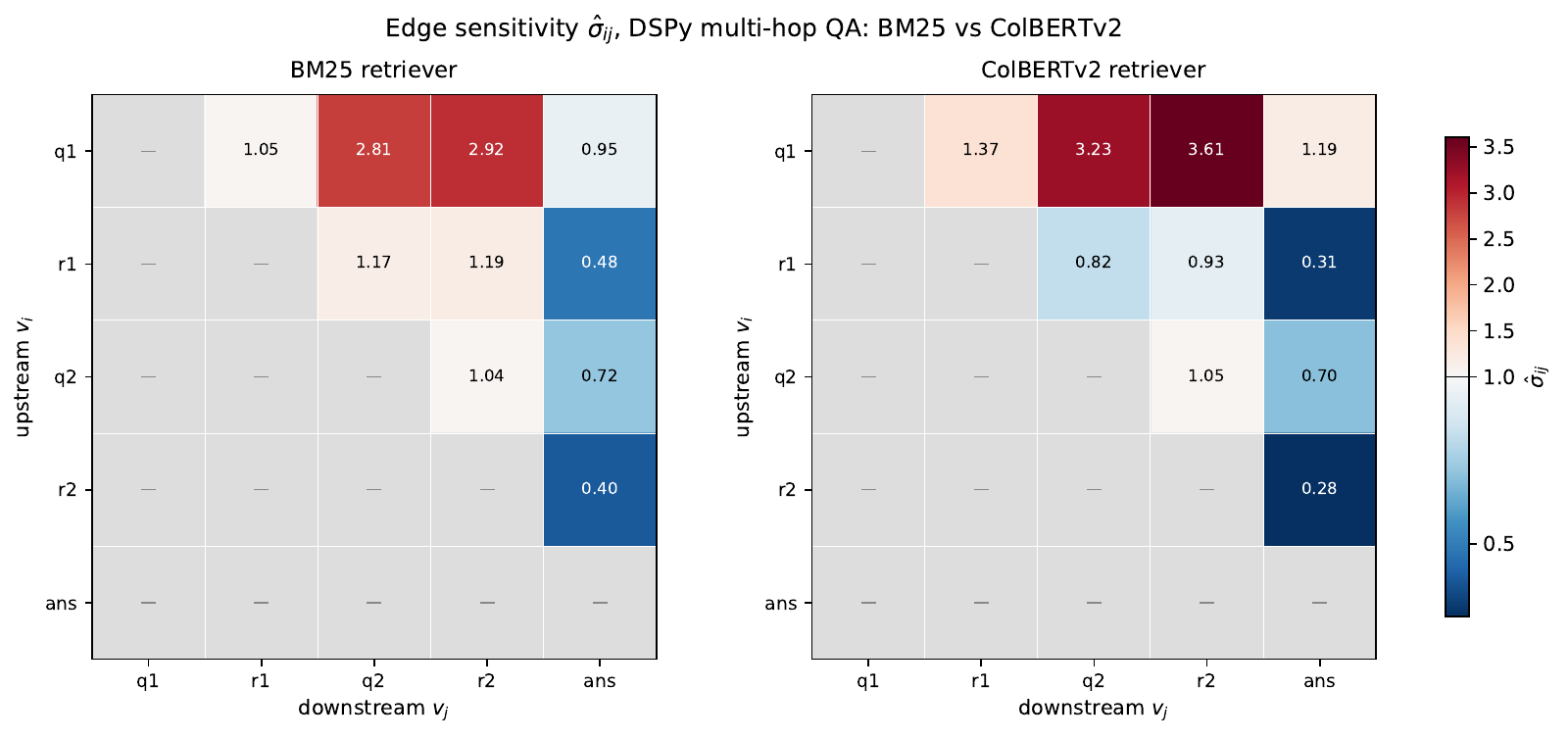}
\caption{\textbf{$\hat{\sigma}$ heatmap, BM25 vs ColBERTv2.} Rows: upstream $v_i$; columns: downstream $v_j$. Color encodes $\hat{\sigma}_{ij}$ on a diverging scale centered at $\hat{\sigma}{=}1$ (white). Greyed cells are temporally infeasible (i fires after j) or have insufficient data ($n(d_i{>}\epsilon){<}30$). Strong amplifier and absorber cells (deep red, deep blue) are stable across retrievers; only near-unity cells shift color --- the same three edges marked \emph{flip} in Table~\ref{tab:sigma_dspy_compare}.}
\label{fig:sigma_heatmap_dspy}
\end{figure}

Table~\ref{tab:sigma_dspy_compare} reports $\hat{\sigma}$ and its decomposition (median ratio, intrinsic $d_j$ when $d_i\!\approx\!0$) for every observed temporally-ordered edge in the DSPy multi-hop QA pipeline (Section~\ref{sec:dspy}) under both retrievers: ColBERTv2 and BM25, both at 500 dev questions $\times$ 3 repeats $=$ 1{,}500 pairs (matched protocol; same pipeline and determinism contract). Figure~\ref{fig:sigma_heatmap_dspy} plots the same data as side-by-side heatmaps for visual comparison.

The strong amplifiers ($\hat{\sigma}{>}2$) and strong absorbers ($\hat{\sigma}{<}0.5$) retain their sign and magnitude character across retrievers. Three near-unity edges ($|\hat{\sigma}-1|<0.4$) cross $\hat{\sigma}=1$ between retrievers (marked \emph{flip}); these sit precisely in Definition~3's most decision-sensitive regime. The decomposition isolates the mechanism even where classification agrees: on \textsc{gen\_query\_1}$\to$\textsc{retrieve\_1} (an amplifier under both retrievers, $\hat{\sigma}{=}1.05$ vs $1.37$), the intrinsic-noise term $d_{\text{retrieve\_1}}$ in pairs where $d_{\text{gen\_query\_1}}\!\approx\!0$ is exactly $0.000$ under \emph{both} retrievers, while the median ratio $d_{\text{retrieve\_1}}/d_{\text{gen\_query\_1}}$ jumps from $0.26$ (BM25) to $0.86$ (ColBERTv2) --- upstream-coupling sensitivity, not per-node noise.

\begin{table}[h]
\caption{\textbf{Cross-retriever $\hat{\sigma}$ decomposition.} For each ordered edge, $\hat{\sigma}$, median ratio $d_j/d_i$ in drifted pairs, and intrinsic $d_j$ (mean $d_j$ in pairs where $d_i\!\approx\!0$) under both retrievers. Bold $\hat{\sigma}$ values are amplifiers ($\hat{\sigma}{>}1$). Edges marked \emph{flip} cross $\hat{\sigma}{=}1$ between retrievers; all are near-unity. $^*$transitive (no direct edge).}
\label{tab:sigma_dspy_compare}
\centering
\footnotesize
\setlength{\tabcolsep}{4pt}
\begin{tabular}{lrrrrrrc}
\toprule
& \multicolumn{3}{c}{BM25 (1{,}500 pairs)} & \multicolumn{3}{c}{ColBERTv2 (1{,}500 pairs)} & \\
\cmidrule(lr){2-4} \cmidrule(lr){5-7}
edge ($v_i \to v_j$) & $\hat{\sigma}$ & median & intr.\ $d_j$ & $\hat{\sigma}$ & median & intr.\ $d_j$ & flip? \\
\midrule
generate\_query\_1 $\to$ generate\_query\_2 & \textbf{2.805} & 1.193 & 0.267 & \textbf{3.230} & 1.182 & 0.280 & --- \\
generate\_query\_1 $\to$ retrieve\_2$^*$    & \textbf{2.919} & 0.987 & 0.283 & \textbf{3.606} & 1.041 & 0.341 & --- \\
generate\_query\_1 $\to$ retrieve\_1        & \textbf{1.049} & 0.258 & 0.000 & \textbf{1.369} & 0.864 & 0.000 & --- \\
generate\_query\_1 $\to$ generate\_answer$^*$ & 0.945        & 0.267 & 0.094 & \textbf{1.189} & 0.269 & 0.073 & \emph{flip} \\
generate\_query\_2 $\to$ retrieve\_2        & 1.042          & 0.792 & 0.000 & 1.049          & 0.918 & 0.000 & --- \\
retrieve\_1        $\to$ generate\_query\_2 & \textbf{1.175} & 0.764 & 0.273 & 0.823          & 0.716 & 0.257 & \emph{flip} \\
retrieve\_1        $\to$ retrieve\_2        & \textbf{1.195} & 0.944 & 0.272 & 0.928          & 0.886 & 0.277 & \emph{flip} \\
generate\_query\_2 $\to$ generate\_answer   & 0.722          & 0.240 & 0.036 & 0.698          & 0.213 & 0.097 & --- \\
retrieve\_1        $\to$ generate\_answer   & 0.479          & 0.177 & 0.107 & 0.311          & 0.160 & 0.096 & --- \\
retrieve\_2        $\to$ generate\_answer   & 0.399          & 0.167 & 0.083 & 0.282          & 0.146 & 0.083 & --- \\
\bottomrule
\end{tabular}
\end{table}

\begin{figure}[h]
\centering
\includegraphics[width=\textwidth]{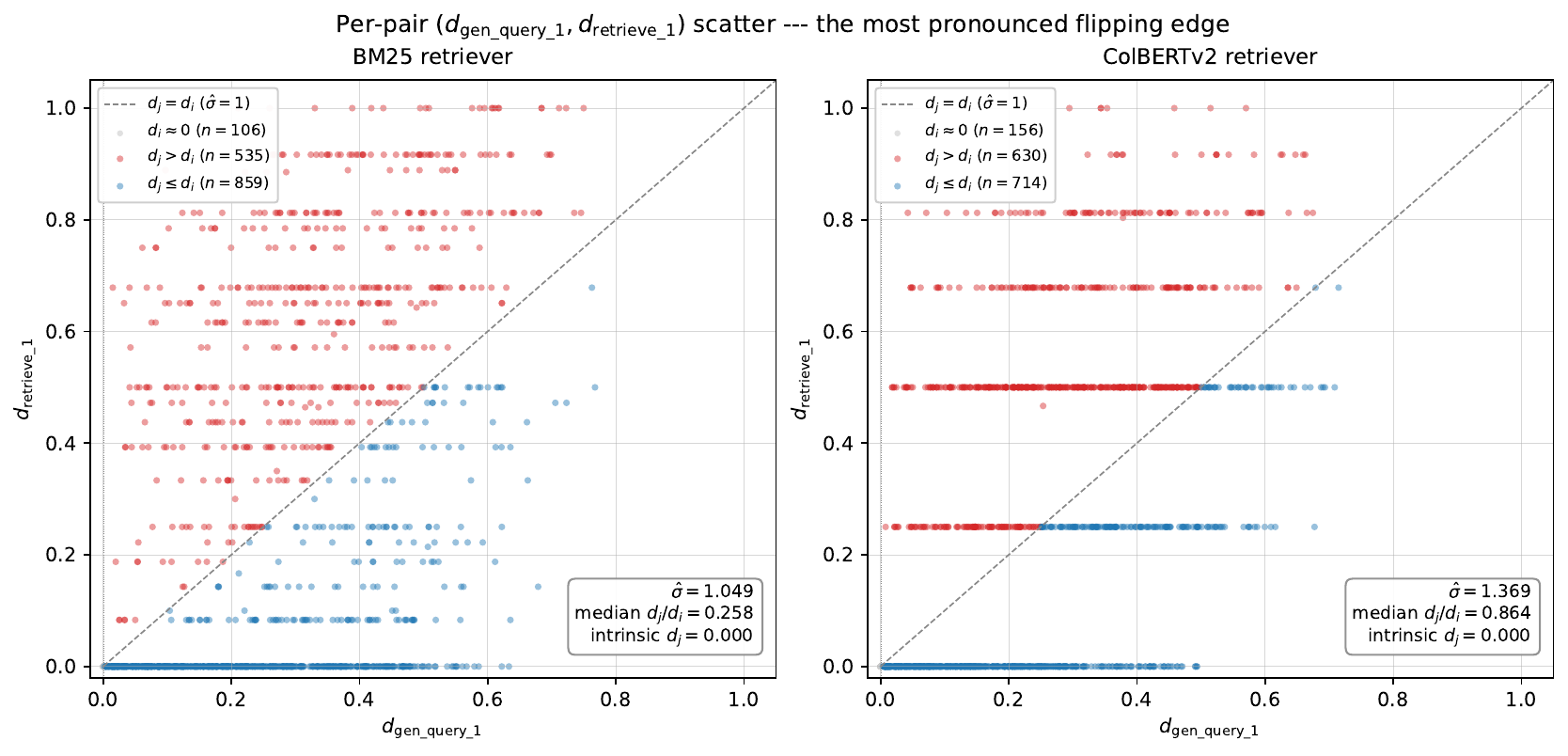}
\caption{\textbf{Per-pair $(d_i, d_j)$ scatter for the most pronounced flipping edge.}
Each dot is one same-input pair. Dashed diagonal is $d_j{=}d_i$ ($\hat{\sigma}{=}1$); pairs above amplify, pairs below absorb. Grey dots on the y-axis ($d_i{<}\epsilon$) carry the intrinsic-noise term --- both retrievers cluster these tightly at $d_j\!=\!0$, so the intrinsic noise is exactly $0.000$ in both cases. The cloud of drifted pairs ($d_i{>}\epsilon$) sits well below the diagonal under BM25 (median ratio $0.26$) but spreads diagonally under ColBERTv2 (median ratio $0.86$) --- the framework's decomposition reads this directly as upstream-coupling sensitivity, not increased per-node noise.}
\label{fig:sigma_scatter_dspy}
\end{figure}

\section{Case Studies: From Framework Outputs to Production Changes}
\label{app:case_studies}

This appendix documents four production interventions on the systems studied in this paper, illustrating how QUIVER's pipeline-level signals become actionable for practitioners. Three were directly driven by framework analysis (Sections~\ref{app:case_p_planner}, \ref{app:case_q_reranker}, and \ref{app:case_p_faithfulness}); one (Section~\ref{app:case_p_rewriter}) predates the framework's formalism and is included because the patterns the framework now articulates were the operational signals that motivated its development. Each case lists the framework finding, the operational response, the observed outcome, and what component-level evaluation alone would have missed.

\subsection{System P: Rewriter Variant Diversification}
\label{app:case_p_rewriter}

\noindent\textit{Background.} QUIVER's design was motivated in part by operational debugging on System P, where pipeline-level coupling --- small rewriter changes producing disproportionate downstream drift and triggering unnecessary loop iterations --- was an observable but unformalized pain point that node-level evaluation of the rewriter could not surface.

\noindent\textit{Operational response (predates framework).} The team's intervention, implemented before this paper's $\hat{\sigma}$ formalism existed, was to modify the rewriter to produce \textbf{multiple semantically distinct rewrites} in parallel, with deduplication on the retrieved-fragment set. The intuition was that a portfolio of rewrites would average out the variance any single rewrite contributed.

\noindent\textit{Retrospective framework analysis.} QUIVER's $\hat{\sigma}$ matrix on the System P observational corpus (collected on the diversified-rewriter system) shows the rewriter retains a load-bearing, threshold-sensitive profile: $\hat{\sigma}(\text{rewriter}\to\text{discovery})=1.128$ with a bimodal ratio distribution (most pairs absorb, a minority tail amplifies). Under interventional rewriter perturbation, $D_{\mathrm{shape}}$ rises from 12.9\% (observational) to 37.5\% (interventional) and $D_{\mathrm{iter}}$ from 3.3\% to 9.5\% --- single-prompt perturbation still bifurcates the execution path in roughly one in three pairs. The framework provides formal vocabulary for what the team had observed informally: rewriter variance couples upstream of a structural bifurcation, and diversification reduces single-point coupling.

\noindent\textit{What this case study illustrates.} Component-level evaluation of the rewriter would have shown rewrites of acceptable quality. The pipeline-level signal --- that rewriter variance bifurcates downstream execution paths --- is invisible to isolated benchmarks. The framework's $\hat{\sigma}$ classifies the edge, the $D_{\mathrm{shape}}$ rates quantify the bifurcation, and the bimodal threshold-sensitive pattern is now a well-defined diagnostic primitive applicable to other pipelines.

\subsection{System P: Resolving a Categorical-Flip Amplifier at Planner $\to$ Composer}
\label{app:case_p_planner}

\noindent\textit{Framework finding.} On System P's observational corpus, $\hat{\sigma}(\text{planner}\to\text{composer})=1.069$ with median ratio $0.537$ --- a bimodal amplifier where the majority of same-input pairs absorb (median well below mean) but a minority tail amplifies sharply. The planner was also the \textbf{sole bifurcation source} in the pipeline: $\hat{\beta}_{\mathrm{shape}}=0.102$ at planner, $0$ at every other upstream node. Together these signals identified planner $\to$ composer as a categorical-flip amplifier: small planner output drift could push a routing-field categorical across its decision boundary, triggering a structurally different downstream composition path with end-user-visible irrecoverable consequences.

\noindent\textit{Operational response (driven by framework analysis).} The framework's diagnosis pointed to two complementary interventions. \textbf{Architecturally}, the team introduced a \textbf{hybrid composer variant} for the boundary case --- a path that absorbs the categorical variance into a single, robust output rather than propagating it as an irrecoverable structural divergence. \textbf{Operationally}, the planner's prompt was tuned to favor the boundary-case category more aggressively, reducing the rate at which a single planner-output difference triggered diametrically opposed composer behaviors.

\noindent\textit{Outcome.} The categorical-flip mechanism was dampened: same-input pairs that previously routed through entirely different downstream paths more often share the hybrid path. End-user-visible irrecoverable differences across same-input runs were reduced.

\noindent\textit{What this case study illustrates.} Component-level evaluation of the planner (e.g., intent-classification accuracy on a curated benchmark) would have shown a node performing its assigned task. The pipeline-level signal --- that small classification flips at a structural boundary irrecoverably altered user-visible behavior --- is exactly what $\hat{\sigma}$'s bimodal-amplifier pattern (mean above 1, median well below) and $\hat{\beta}_{\mathrm{shape}}$'s bifurcation lower bound are designed to surface. The framework flagged the problem; the architectural + prompt-level intervention addressed both its magnitude (the amplifier) and its structural mechanism (the bifurcation).

\subsection{System Q: Reranker Removal}
\label{app:case_q_reranker}

\noindent\textit{Framework finding.} On the System Q observational corpus, the Reranker had the highest $D_{\mathrm{noise}}$ of any node ($\approx 2\times$ the next-noisiest stage) while the downstream Reranker $\to$ Generator edge was a strong absorber ($\hat{\sigma}=0.341$, median ratio $0.303$). Upstream stages amplified heavily into it ($\hat{\sigma}(\text{Fast LLM}\to\text{Reranker})=9.018$, $\hat{\sigma}(\text{Slow LLM}\to\text{Reranker})=5.982$). The framework's diagnostic profile: a high-noise node feeding a leaky absorber --- a pattern in which the node consumes the lion's share of pipeline variance, with no commensurate downstream value evident.

\noindent\textit{Operational response (driven by framework analysis).} The pattern flagged the Reranker as a candidate for ablation. The team ran the ablation, removing the Reranker from the pipeline; the ablation confirmed the framework's hypothesis: removing the Reranker did not regress quality.

\noindent\textit{Outcome.} The Reranker was fully removed from the production pipeline. Generator output quality improved, with reduced variance and drift across same-input runs; end-to-end latency improved (one fewer model call per query).

\noindent\textit{What this case study illustrates.} Component-level evaluation of the Reranker (rerank-quality benchmarks, MRR, NDCG on curated test sets) would have shown a node that performs its assigned task. The pipeline-level signal --- that this node was the dominant variance source whose downstream contribution did not justify its instability --- is a \emph{coupling} property no isolated benchmark surfaces. QUIVER's $D_{\mathrm{noise}}$ + $\hat{\sigma}$ pairing made the variance attribution explicit; ablation confirmed; the production change followed.

\subsection{System P: Stale Golden Data Detected by Distribution Faithfulness}
\label{app:case_p_faithfulness}

\noindent\textit{Framework finding.} Computing the distribution faithfulness gap (Principle~1) on System P surfaced fragment-related golden fields with anomalously high gaps across three downstream nodes (discovery, planner, composer; see Table~\ref{tab:faithful}). The pattern was localized to fragment-reference fields rather than spread across all field types, suggesting a structural mismatch in evaluation-data coverage rather than model misalignment.

\noindent\textit{Operational diagnosis.} Investigation traced the elevated gaps to a configuration change that had increased retrieval depth from 10 to 20 fragments without a corresponding update to the golden evaluation set; the golden expectations now covered at most half the current retrieval set by construction. End-to-end evaluation did not detect this drift, and per-node evaluation would have attributed the resulting score degradation to the nodes themselves rather than to stale evaluation data.

\noindent\textit{Outcome.} The golden dataset was regenerated against the current retrieval depth, restoring per-node evaluation validity for the affected fields.

\noindent\textit{What this case study illustrates.} Distribution faithfulness measurement at the per-node, per-field level surfaces evaluation invalidation that aggregate metrics cannot localize. A single configuration change can silently invalidate evaluation across multiple downstream nodes; the framework localizes the cause to specific field categories, distinguishing stale-evaluation artifacts from genuine model drift.